\documentclass[conference,10pt]{IEEEtran}
\IEEEoverridecommandlockouts
\usepackage{cite}
\usepackage{amsmath,amssymb,amsfonts}
\usepackage{graphicx}
\usepackage{textcomp}
\usepackage{amsmath}
\usepackage{booktabs} 
\usepackage{multirow}
\usepackage[T1]{fontenc}
\usepackage{stfloats}

\usepackage[utf8]{inputenc}
\usepackage{hyperref}       
\usepackage{url}            
\usepackage{booktabs}       
\usepackage{amsfonts}       
\usepackage{nicefrac}       
\usepackage{microtype}      
\usepackage{xcolor}         
\usepackage{algorithm}

\usepackage{algpseudocode}
\usepackage{amsmath}
\usepackage{tikz}
\usepackage[skip=0pt]{caption}

\usepackage{orcidlink}
\usepackage{comment}
\usepackage{xcolor}
\usepackage{colortbl}
\usepackage{svg}
\usepackage{booktabs}
\usepackage{tabularx} 
\usepackage{ragged2e}
\usepackage{amssymb}
\usepackage{braket}
\usepackage{xcolor}
\def\BibTeX{{\rm B\kern-.05em{\sc i\kern-.025em b}\kern-.08em
    T\kern-.1667em\lower.7ex\hbox{E}\kern-.125emX}}
\usepackage{eurosym}
\definecolor{good}{RGB}{230,245,233}   
\definecolor{mid}{RGB}{255,249,224}    
\definecolor{bad}{RGB}{252,232,232}    

\definecolor{headgray}{gray}{0.92}
\definecolor{rowgray}{gray}{0.97}
\definecolor{qganrow}{RGB}{232,242,255}
\definecolor{headgray}{gray}{0.92}
\definecolor{rowgray}{gray}{0.97}
\definecolor{softblue}{RGB}{230,242,255} 
\newcommand{\hb}[1]{\cellcolor{softblue}\textbf{#1}} 
\newcommand{\hcell}[1]{\cellcolor{softblue}#1}      %
\definecolor{headgray}{gray}{0.92}
\definecolor{rowgray}{gray}{0.97}
\definecolor{softblue}{RGB}{230,242,255}

\newcommand{\grayrow}{%
& \cellcolor{rowgray}%
}

\begin{document}

\title{Q-SYNTH: Hybrid Quantum-Classical Adversarial Augmentation for Imbalanced Fraud Detection}

\author{\IEEEauthorblockN{Adam Innan$^{*}$\thanks{$^{*}$Corresponding author}\textsuperscript{1}, Mansour El Alami\textsuperscript{1},  Nouhaila Innan\orcidlink{0000-0002-1014-3457}\textsuperscript{2,3}, Muhammad Shafique\orcidlink{0000-0002-2607-8135}\textsuperscript{2,3}, and Mohamed Bennai\orcidlink{0000-0002-7364-5171}\textsuperscript{1}} 
\IEEEauthorblockA{
\textsuperscript{1}Quantum Physics and Spintronics Team, LPMC, Faculty of Sciences Ben M'sick,\\ Hassan II University of Casablanca, Morocco\\
\textsuperscript{2}eBRAIN Lab, Division of Engineering, New York University Abu Dhabi (NYUAD), Abu Dhabi, UAE\\
\textsuperscript{3}Center for Quantum and Topological Systems (CQTS), NYUAD Research Institute, NYUAD, Abu Dhabi, UAE\\
adam.innan-etu@etu.univh2c.ma, mansour.elalami-etu@etu.univh2c.ma, nouhaila.innan@nyu.edu,\\ muhammad.shafique@nyu.edu, mohamed.bennai@univh2c.ma\\
}}

\maketitle

\begin{abstract}
Credit card fraud detection is fundamentally challenged by extreme class imbalance, where fraudulent transactions are rare yet operationally critical. This imbalance often biases supervised learners toward the legitimate class, leading to high overall accuracy but weaker fraud-class recall and F1-score. This paper introduces Q-SYNTH, a hybrid classical--quantum generative adversarial framework in which a parameterized quantum circuit serves as the generator and a classical neural network serves as the discriminator. Q-SYNTH is designed for minority-class fraud synthesis in tabular data and is evaluated along two dimensions: statistical fidelity to real fraud samples and downstream performance for fraud detection. To this end, generated samples are assessed using distributional similarity measures based on Kolmogorov--Smirnov statistics and Wasserstein distances, real-vs-synthetic detectability measured by AUC-ROC, and downstream classification performance across both quantum and classical classifiers. Under the reported protocol, Q-SYNTH reduces marginal distribution mismatch relative to a classical GAN baseline while maintaining competitive downstream fraud-detection performance. Although SMOTE achieves the strongest feature-wise similarity and the classical GAN attains the highest downstream performance in several settings, Q-SYNTH offers a favorable compromise between distributional fidelity and downstream performance, supporting the feasibility of hybrid quantum augmentation for imbalanced fraud detection.
\end{abstract}


\section{Introduction}
The rapid expansion of digital payments has increased both the scale and the sophistication of fraud, turning fraud detection into a persistent operational and security challenge for financial institutions and consumers. Recent industry reporting highlights that scam activity continues to accelerate: BioCatch's 2025 analysis \cite{Banking}, based on financial institutions serving nearly 350 million consumers across five continents, reports a 65\% year-over-year increase in scam attempts, including a 100\% spike in voice phishing (vishing), a 63\% increase in romance scam attempts, and a 42\% increase in attempted investment scams, alongside a tenfold increase in SMS phishing activity. In parallel, organizational exposure remains high: the 2025 AFP Payments Fraud and Control Survey reports that 79\% of organizations experienced payments fraud attacks or attempts in 2024 \cite{bank2}, with business email compromise (BEC) cited as the leading avenue for fraud attempts by 63\% of respondents. For the UK specifically \cite{BBC_2025}, UK Finance reports that in the first half of 2025, criminals stole \pounds 629.3 million through scams and payment fraud, with over 2 million reported cases during January--June 2025. These trends reinforce a central reality: fraud is not only prevalent but also adaptive, and detection systems must cope with continuously evolving attack patterns.

A key technical barrier in fraud detection is the extreme class imbalance inherent to transactional datasets, where fraudulent events constitute a tiny fraction of all records \cite{chen2025deep,hafez2025systematic}. Under such imbalance, standard supervised learners can attain deceptively high overall accuracy by prioritizing the majority class while failing to recover rare fraud patterns, typically resulting in low fraud-class recall and fraud-class F1-score. Beyond scarcity, the minority class often exhibits heterogeneous and evolving modes, making it difficult for models to generalize from limited examples to novel fraud behaviors. For this reason, synthetic data generation and augmentation have become practical tools for improving minority-class learning: by enriching training data with additional fraud-like samples, augmentation increases the effective support of the minority distribution, reduces overfitting to a small number of fraud instances, and can shift the learned decision boundary toward improved detection of rare events.

However, existing augmentation approaches face a well-known tension between distributional fidelity and downstream performance. Interpolation-based oversampling methods such as SMOTE often preserve local feature statistics and can yield strong marginal similarity to real fraud samples, but they may introduce limited diversity, remain confined to convex combinations, and yield smaller gains in decision-relevant regions. Deep generative models, including GAN variants, can introduce more diverse samples that benefit classification, yet they may fail to match the fine-grained statistics of real minority-class distributions in highly imbalanced tabular settings, raising concerns about distributional fidelity, real-vs-synthetic detectability, and stability. 

Importantly, the fraud literature often emphasizes downstream classification metrics without systematically validating the similarity between real and synthetic fraud distributions \cite{tayebi2025generative,nama2025credit,malhotra2025credit,kk2026mitigating}, which limits interpretability and weakens claims about sample realism. Although some studies report a combination of similarity and classification metrics, such evidence remains sparse and heterogeneous across datasets and evaluation protocols. This lack of consistent similarity auditing is particularly limiting when synthetic samples are intended for operational or security-sensitive settings.

In parallel, Quantum Machine Learning (QML) has been increasingly explored for fraud detection \cite{grossi2022mixed,innan2024financial1,innan2024financial,innan2025qfnn,el2026comparative,sawaika2025privacy,innan2025circuithunt,alami2025fid}, but the dominant focus has been on quantum classifiers, such as Quantum Neural Networks (QNNs), rather than on quantum-enhanced data generation. This leaves an important open question: \textit{can a quantum generator serve as a practical augmentation mechanism for minority-class tabular data while maintaining measurable statistical fidelity to real fraud samples?} Addressing this question is timely for two reasons. First, generative augmentation is directly aligned with the central bottleneck in fraud detection, namely, the limited availability and diversity of fraud examples. Second, parameterized quantum circuits provide a distinct modeling class for hybrid generative learning, making them a natural candidate for investigating minority-class synthesis in low-dimensional tabular representations.

\begin{figure}[htpb]
    \centering
    \includegraphics[width=1\linewidth]{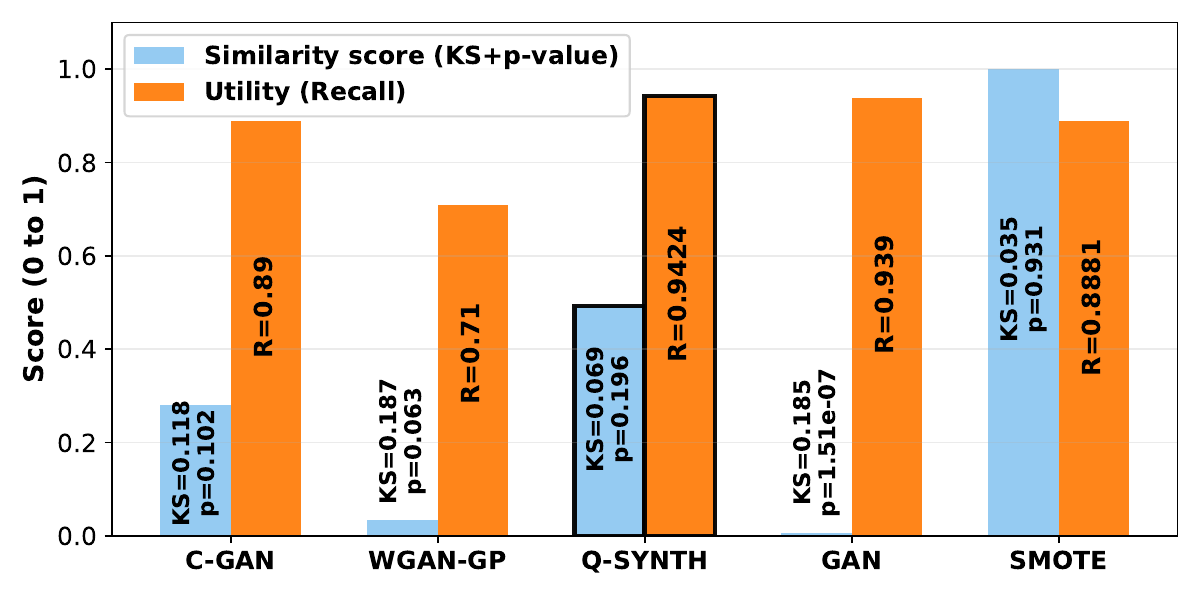}
    \caption{\small Motivation summary comparing distribution similarity and classification performance for five augmentation methods: C-GAN \cite{hasan2025generative}, WGAN-GP \cite{gulrajani2017improved}, GAN, SMOTE, and Q-SYNTH. The Similarity score (blue) is a single value in [0,1] formed by averaging two normalized signals: an inverted KS statistic (lower KS is better) and the KS $p$-value (higher is better). The Utility score (orange) is the downstream fraud Recall $R$ in [0,1]. The figure is intended only as an intuitive summary of the fidelity--performance trade-off; all substantive comparisons in the paper rely on the raw metrics reported in the tables.}
    \label{histogram}
\end{figure}

To this end, we propose \textbf{Q-SYNTH}, a hybrid quantum--classical adversarial framework for tabular fraud augmentation in which a parameterized quantum circuit serves as the generator and a classical network serves as the discriminator. Q-SYNTH is evaluated under a unified protocol that assesses both distributional fidelity and downstream fraud-detection performance. Under the reported protocol, Q-SYNTH reduces the marginal-similarity gap relative to a classical GAN baseline while maintaining competitive downstream performance across quantum and classical classifiers. Taken together, these results suggest that Q-SYNTH offers a favorable trade-off between statistical fidelity and downstream performance, as illustrated qualitatively in Fig.~\ref{histogram}.

The main contributions of this work are as follows:
\begin{itemize}
  \item \textbf{Hybrid quantum--classical fraud augmentation:} We introduce Q-SYNTH, a hybrid adversarial framework in which a variational quantum generator synthesizes minority-class fraud samples and a classical discriminator guides training in a tabular setting.
  \item \textbf{Controlled synthesis pipeline for tabular fraud data:} We develop an end-to-end workflow from preprocessing and bounded representation mapping to inverse transformation back to the feature space used for downstream fraud detection.
  \item \textbf{Joint evaluation of distributional fidelity and downstream performance:} We assess generated samples using distributional similarity measures, real-vs-synthetic detectability, and downstream fraud-detection performance under a unified evaluation protocol.
  \item \textbf{Empirical trade-off analysis across learners:} Under the reported protocol, Q-SYNTH improves marginal fidelity relative to a classical GAN baseline while maintaining competitive downstream performance across quantum and classical classifiers.
\end{itemize}

The rest of the paper is organized as follows. Sec.~\ref{sec2} reviews the fraud-detection setting and prior augmentation approaches, motivating hybrid QGAN-based synthesis. Sec.~\ref{sec3} details the proposed \textit{Q-SYNTH} pipeline, including preprocessing, stabilized adversarial training, and the evaluation protocol. Sec.~\ref{sec4} reports experimental results on training dynamics, fidelity, downstream detection performance, and scaling with the synthetic-fraud ratio. Sec.~\ref{sec5} concludes and outlines future directions.
\section{Background and Related Work}\label{sec2}

\subsection{Fraud detection and the class-imbalance bottleneck}
Credit card fraud detection remains a core challenge for payment systems because fraudulent transactions incur significant losses while occurring rarely compared to legitimate activity. Real-world fraud datasets are typically extremely imbalanced, with fraud often accounting for well below 1\% of all transactions~\cite{xu2022classifier}. This skew biases supervised learners toward the majority class and can yield misleadingly high overall accuracy while missing the minority class that matters most, leading to degraded fraud-class recall and F1-score across common classifiers (e.g., logistic regression, SVMs, random forests, and neural networks)~\cite{feng2018learning, alarfaj2022credit, mienye2024deep, gaav2025recent}. As a result, mitigating imbalance is not a secondary preprocessing step but a central design requirement for robust fraud detection.

\subsection{Classical resampling: strong local fidelity, limited minority coverage}
A broad family of classical remedies relies on resampling. Popular oversampling methods such as SMOTE~\cite{chawla2002smote}, Borderline-SMOTE~\cite{han2005borderline}, and ADASYN~\cite{he2008adasyn} generate synthetic minority samples by interpolating between nearby fraud instances in feature space. These techniques often preserve local, feature-wise statistics and can improve training stability in low-data regimes. However, they are fundamentally constrained by their interpolation mechanism: they may fail to capture complex, multi-modal minority distributions, can produce unrealistic samples in high-dimensional spaces, and may not enrich decision-relevant regions near difficult class boundaries~\cite{hong2024time}. In practice, this can create a mismatch between marginal similarity and downstream performance, where samples that appear statistically close to the minority class do not necessarily induce the strongest improvements in fraud-class recall and F1-score. This trade-off motivates generative approaches that learn the minority distribution rather than interpolating it.

\subsection{GAN-based fraud augmentation: promising downstream performance, but evaluation is often classification-focused}
GANs provide a more expressive alternative by learning a data distribution via an adversarial game between a generator and a discriminator~\cite{goodfellow2014generative}. In fraud detection, multiple studies use GAN-generated fraud samples as augmentation to improve minority-class detection, reporting notable gains compared to training on the original imbalanced data and, in many cases, compared to classical oversampling~\cite{fiore2019using, hwang2020efficient, strelcenia2022comparative, yang2021ida, sharma2022smotified, ali2024improving, james2024adversarial, zhu2024utilizing, strelcenia2023new, hasan2025generative, kumar2025fraud, ke2025detection, mohsen2025robust, tayebi2025generative, wang2025gan_bert, zheng2025integrated}. These works suggest that adversarial augmentation can inject task-relevant variability into the minority class and improve fraud-class sensitivity.

Despite these successes, two limitations repeatedly appear in the fraud-augmentation literature. First, training instability (oscillations, mode collapse) and hyperparameter sensitivity remain common, especially when the minority class is scarce, which can amplify noise in the augmented training set and introduce learning errors~\cite{ghaleb2023ensemble}. Second, many fraud-focused GAN studies emphasize downstream classification metrics, while distributional fidelity and detectability of generated samples (real-vs-synthetic detectability) are reported less consistently and under non-unified protocols, making it difficult to compare methods on a common fidelity--performance basis. This motivates an evaluation that jointly reports distributional similarity and downstream performance under a controlled protocol.

\subsection{Quantum GANs: foundations and motivation for hybrid generators}
With the emergence of QML~\cite{biamonte2017quantum,schuld2015introduction,rodriguez2025survey}, Quantum Generative Adversarial Networks (QGANs) have been proposed as quantum analogs of adversarial training, where the generator and/or discriminator can be implemented with quantum information processors~\cite{lloyd2018quantum}. Foundational work establishes that the adversarial game admits a fixed point where generated statistics match the data and the discriminator cannot distinguish real from generated samples, and argues that quantum adversarial models may be particularly relevant when data arises from measurements in high-dimensional spaces~\cite{lloyd2018quantum}. Subsequent studies demonstrate practical circuit constructions and gradient computation for training QGANs, supporting the feasibility of adversarial learning with parameterized quantum circuits~\cite{dallaire2018quantum}. Hybrid formulations, where a quantum generator interacts with a classical discriminator, have also been explored for learning and loading probability distributions into quantum states efficiently, enabling downstream quantum routines such as amplitude estimation~\cite{zoufal2019quantum}. Beyond tabular distributions, QGANs have been investigated for financial time-series generation and for broader generative tasks, including image and domain-translation settings, highlighting their use across diverse generative applications and motivating further study in structured tabular domains~\cite{dechant2026quantum, yang2026ihqgan, pajuhanfard2024survey, nandal2025image, andra2025data}.

These results motivate hybrid QGANs as a promising direction for tabular minority synthesis: a parameterized quantum generator can serve as a compact distribution model, while a classical discriminator provides a flexible learning signal and a stable scoring interface for real-valued tabular data.

\subsection{Gap and positioning of Q-SYNTH}
While classical GANs are widely used for fraud augmentation, and QGANs have been studied in several domains, including finance-oriented generative modeling, the intersection remains underexplored. Existing quantum-fraud studies primarily emphasize quantum classifiers (e.g., QSVMs and QNNs), whereas QGAN-based synthesis for fraud-imbalance mitigation has received limited attention. Moreover, when GAN-based augmentation is considered in fraud settings, methods are often assessed mainly through classification outcomes rather than through a joint protocol that also examines distributional fidelity and external detectability. This gap is particularly important for fraud detection, where synthetic data must be both statistically faithful, to avoid artifacts and unrealistic patterns, and decision-relevant, to improve fraud-class recall and F1-score under imbalance.

Accordingly, this work introduces \textit{Q-SYNTH}, a hybrid classical--quantum QGAN framework for synthesizing fraudulent transactions, and evaluates it using a unified protocol that reports both distributional similarity (e.g., KS and Wasserstein metrics) and external detectability measured by AUC-ROC, alongside downstream fraud-detection performance across quantum and classical classifiers. This positioning aims to clarify the trade-off between distributional fidelity and downstream performance in fraud augmentation, and to assess the role of a quantum generator within a controlled evaluation pipeline.
\section{Methodology}\label{sec3}
\begin{figure*}[htbp]
    \centering
\includegraphics[width=1\linewidth]{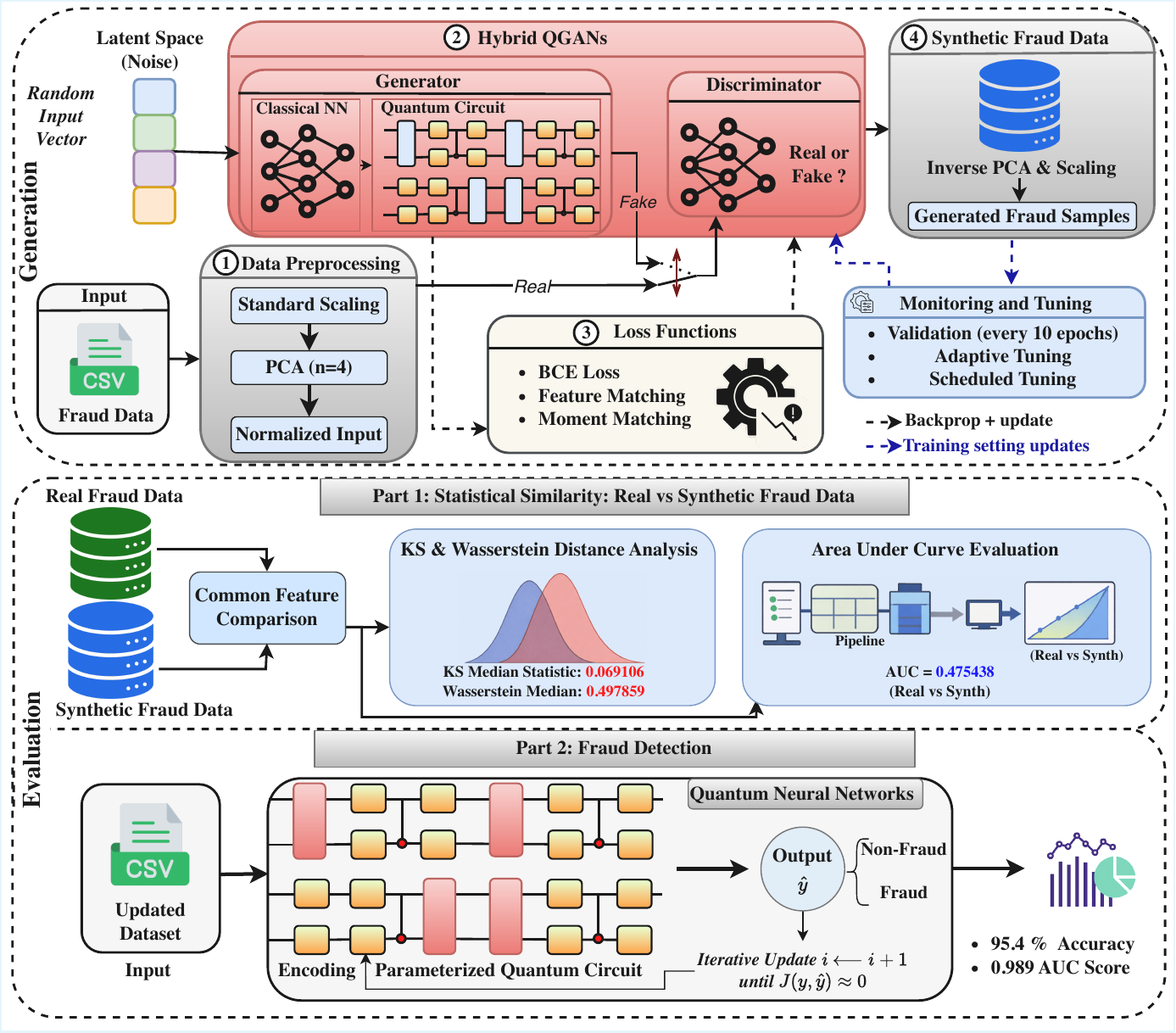}
    \caption{Overview of the proposed Q-SYNTH pipeline for fraud data generation and evaluation.}
    \label{fig:updqgans}
\end{figure*}

This section presents the proposed Q-SYNTH framework for fraud-data generation and evaluation. As summarized in Fig.~\ref{fig:updqgans}, fraud transactions are first transformed into a bounded representation space, after which a hybrid QGAN is trained in that space using a classical network to parameterize a variational quantum circuit and a classical discriminator to distinguish real from generated samples. Training is stabilized through instance noise and regularized objectives, and the resulting generator is then used to produce synthetic fraud records that are mapped back to the feature space used for downstream learning. The generated samples are subsequently assessed for distributional fidelity, external real-vs.-synthetic detectability, and downstream fraud-detection performance.

\subsection{Data Preprocessing}

The preprocessing pipeline transforms the raw transaction records from the Credit Card Fraud Detection dataset \cite{ULB_2018} into a bounded continuous representation that defines the empirical real-data distribution \(p_{\mathrm{data}}\) used for adversarial training and evaluation. Because the generative model is designed to learn the distribution of fraudulent transactions, preprocessing consists of supervised feature selection on the full labeled dataset, followed by restriction to the fraud class, feature standardization, principal component analysis (PCA), and final normalization. This representation is introduced as a tractable low-dimensional space for hybrid quantum generation, rather than as a claim that the full structure of the original tabular distribution is preserved without approximation.

Let the original dataset be the pair \((\mathbf{X}, y)\), where \(\mathbf{X}\in\mathbb{R}^{N\times D}\) is the feature matrix and \(y\in\{0,1\}^{N}\) is the binary class vector. To retain variables that are most informative for fraud discrimination, we apply \textsc{SelectKBest} to the full labeled dataset and keep the top \(k=10\) features according to a univariate scoring function. This yields the reduced matrix
\begin{equation}
\mathbf{X}^{(\mathrm{sel})}\in\mathbb{R}^{N\times 10}.
\end{equation}

We then retain only the fraudulent transactions, obtaining
\begin{equation}
\mathbf{X}^{(1)}\in\mathbb{R}^{N_1\times 10},\qquad
N_1=\sum_{i=1}^{N}\mathbb{I}[y_i=1].
\end{equation}
Let \(x_{n,j}\) denote the \((n,j)\)-th entry of \(\mathbf{X}^{(1)}\), with \(n=1,\dots,N_1\) and \(j=1,\dots,10\).

The selected fraud samples are standardized feature-wise. Let \(\mu\in\mathbb{R}^{10}\) and \(\sigma\in\mathbb{R}^{10}\) denote the empirical mean and standard deviation of \(\mathbf{X}^{(1)}\), defined component-wise as
\begin{equation}
\begin{aligned}
\mu_j &= \frac{1}{N_1}\sum_{n=1}^{N_1} x_{n,j}, \\
\sigma_j &= \sqrt{\frac{1}{N_1}\sum_{n=1}^{N_1}\bigl(x_{n,j}-\mu_j\bigr)^2}, \\
j &= 1,\dots,10.
\end{aligned}
\end{equation}
The standardized matrix \(\mathbf{X}^{(\mathrm{std})}\in\mathbb{R}^{N_1\times 10}\) is therefore
\begin{equation}
\begin{aligned}
X^{(\mathrm{std})}_{n,j}
&=\frac{x_{n,j}-\mu_j}{\sigma_j}, \\
&\qquad n=1,\dots,N_1,\ \ j=1,\dots,10.
\end{aligned}
\end{equation}

Next, PCA is applied to project the standardized features onto a four-dimensional latent representation. This dimensionality is chosen to make the hybrid quantum generator compatible with a small number of qubits while preserving a compact continuous representation for adversarial learning. Let \(\mathbf{W}\in\mathbb{R}^{10\times 4}\) be the projection matrix whose columns are the four leading eigenvectors of the empirical covariance matrix of \(\mathbf{X}^{(\mathrm{std})}\). The projected data are given by
\begin{equation}
\mathbf{X}^{(\mathrm{pca})}=\mathbf{X}^{(\mathrm{std})}\mathbf{W},
\qquad
\mathbf{X}^{(\mathrm{pca})}\in\mathbb{R}^{N_1\times 4}.
\end{equation}

To obtain bounded inputs, each principal component is scaled by its maximum absolute value over the fraud subset. Defining
\begin{equation}
M_j=\max\!\left(\max_{1\le n\le N_1}\left|X^{(\mathrm{pca})}_{n,j}\right|,\varepsilon\right),
\qquad j=1,\dots,4,
\end{equation}
where \(\varepsilon>0\) prevents division by zero, the normalized matrix \(\mathbf{X}^{(\mathrm{norm})}\in\mathbb{R}^{N_1\times 4}\) is defined by
\begin{equation}
X^{(\mathrm{norm})}_{n,j}=\frac{X^{(\mathrm{pca})}_{n,j}}{M_j},
\qquad n=1,\dots,N_1,\ \ j=1,\dots,4.
\end{equation}
By construction, \(X^{(\mathrm{norm})}_{n,j}\in[-1,1]\) for all \(n\) and \(j\). The resulting samples \(\mathbf{X}^{(\mathrm{norm})}\) constitute the real observations drawn from \(p_{\mathrm{data}}\) and define the bounded representation space in which adversarial training is performed.
\subsection{Hybrid Classical--Quantum QGAN Architecture}

After preprocessing, adversarial learning is performed in the bounded \(d\)-dimensional representation space defined by \(\mathbf{X}^{(\mathrm{norm})}\). The hybrid QGAN consists of a generator that combines a classical parameter network with a \(d\)-qubit variational quantum circuit, and a classical discriminator that maps both real and generated samples to a probability in \((0,1)\).

\subsubsection{Generator}
The generator maps a latent variable \(z\in\mathbb{R}^{m}\) to a synthetic sample \(\hat{x}\in\mathbb{R}^{d}\), where \(m\in\mathbb{N}\) is the latent dimension and \(z\sim p(z)\) is drawn from a fixed prior. The hybrid design uses a classical front-end to transform the latent input into circuit parameters, allowing the quantum component to act as a compact variational generator in the bounded representation space.

A first classical stage produces an intermediate representation \(h\in\mathbb{R}^{r}\), with \(r\in\mathbb{N}\), as
\begin{equation}
h=\tanh(W_1 z+b_1),
\end{equation}
where \(W_1\in\mathbb{R}^{r\times m}\) and \(b_1\in\mathbb{R}^{r}\) are trainable parameters and \(\tanh(\cdot)\) is applied element-wise. A second affine map produces a vector \(a\in\mathbb{R}^{3d}\), reshaped into an angle matrix \(\Theta\in\mathbb{R}^{d\times 3}\),
\begin{equation}
a=W_2 h+b_2,\qquad \Theta=\mathrm{reshape}(a;d,3),
\end{equation}
where \(W_2\in\mathbb{R}^{3d\times r}\) and \(b_2\in\mathbb{R}^{3d}\) are trainable parameters. For each qubit \(q\in\{1,\dots,d\}\), the triple \((\Theta_{q,1},\Theta_{q,2},\Theta_{q,3})\) parameterizes the single-qubit rotations used in the variational block.

\begin{figure*}[htpb]
    \centering
    \includegraphics[width=1\linewidth]{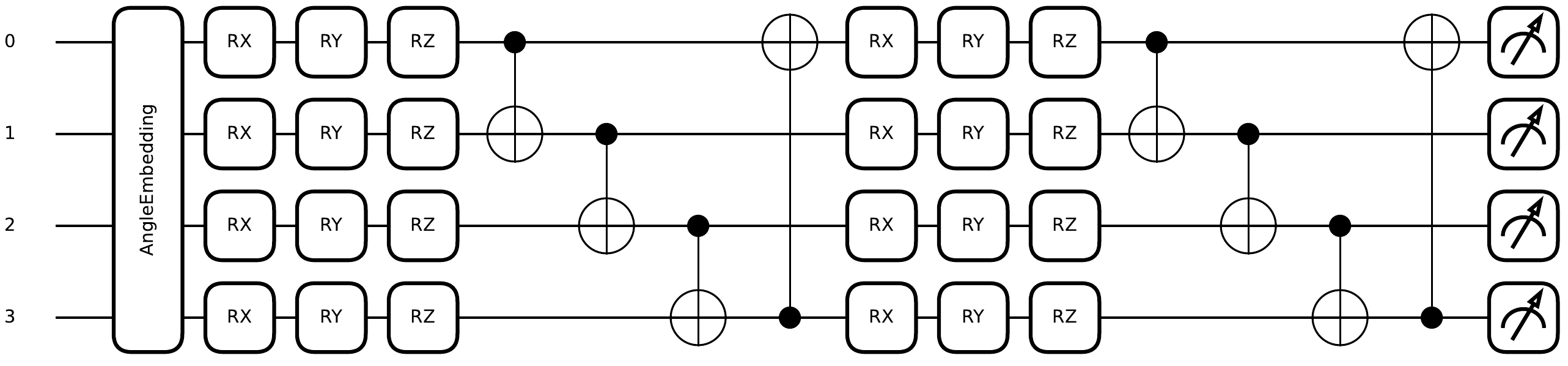}
    \caption{Quantum generator circuit. It starts with angle embedding, followed by trainable single-qubit rotations (RX, RY, RZ) and a ring of CNOT entangling gates. The circuit ends with Pauli-Z measurements on all qubits.}
    \label{circuit}
\end{figure*}

The quantum generator prepares a \(d\)-qubit state from \(\lvert 0\rangle^{\otimes d}\) using an angle-embedding stage followed by \(L\in\mathbb{N}\) variational layers (see Fig.~\ref{circuit}). The prepared state is
\begin{equation}
\lvert \psi(z,\Theta)\rangle = U(z,\Theta)\lvert 0\rangle^{\otimes d}.
\end{equation}

To ensure dimensional consistency with a \(d\)-qubit embedding, define an embedding vector \(z'\in\mathbb{R}^{d}\) obtained from \(z\) through a fixed linear projection followed by element-wise scaling,
\begin{equation}
\begin{aligned}
z'=\pi(z)=S(Pz), \qquad &P\in\mathbb{R}^{d\times m},\\
& S\in\mathbb{R}^{d\times d}\ \text{diagonal},
\end{aligned}
\end{equation}
so that \(z'=(z'_1,\dots,z'_d)^\top\) provides one real-valued embedding angle per qubit.

The embedding unitary applies \(Y\)-rotations on each qubit,
\begin{equation}
U_{\mathrm{enc}}(z')=\prod_{q=1}^{d} R_Y(z'_q),
\qquad
R_Y(\varphi)=\exp\!\left(-i\frac{\varphi}{2}Y\right),
\end{equation}
where \(Y\) is the Pauli-\(Y\) operator. Each variational layer applies three parameterized rotations per qubit followed by ring entanglement. The rotation block is
\begin{equation}
U_{\mathrm{rot}}(\Theta)=\prod_{q=1}^{d} R_X(\Theta_{q,1})\,R_Y(\Theta_{q,2})\,R_Z(\Theta_{q,3}),
\end{equation}
with \(R_X(\varphi)=\exp\!\left(-i\frac{\varphi}{2}X\right)\) and \(R_Z(\varphi)=\exp\!\left(-i\frac{\varphi}{2}Z\right)\), where \(X\) and \(Z\) are the Pauli-\(X\) and Pauli-\(Z\) operators. Entanglement is introduced using a ring of controlled-NOT gates,
\begin{equation}
U_{\mathrm{ent}}=\prod_{q=1}^{d}\mathrm{CNOT}_{q\rightarrow q^+},
\qquad
q^+=\begin{cases}
q+1,& q<d,\\
1,& q=d.
\end{cases}
\end{equation}

In this formulation, the same angle matrix \(\Theta\) is shared across the \(L\) layers as a parameter-efficient choice. Let the \(\ell\)-th layer unitary be \(U_{\ell}(z',\Theta)=U_{\mathrm{ent}}\,U_{\mathrm{rot}}(\Theta)\) for \(\ell\in\{1,\dots,L\}\). The overall state-preparation unitary is
\begin{equation}
U(z,\Theta)=\left(\prod_{\ell=1}^{L}U_{\ell}(z',\Theta)\right)U_{\mathrm{enc}}(z').
\end{equation}

The generator output is obtained by Pauli-\(Z\) expectation-value readout. For each \(q\in\{1,\dots,d\}\),
\begin{equation}
\hat{x}_q=\langle \psi(z,\Theta)\rvert Z_q \lvert \psi(z,\Theta)\rangle,
\end{equation}
where \(Z_q\) acts as Pauli-\(Z\) on qubit \(q\) and identity elsewhere. The synthetic sample is \(\hat{x}=(\hat{x}_1,\dots,\hat{x}_d)\in\mathbb{R}^{d}\). Since \(\hat{x}_q\in[-1,1]\), the generator naturally produces bounded outputs consistent with the preprocessed data domain.

\subsubsection{Discriminator}
The discriminator is a classical mapping \(D:\mathbb{R}^{d}\rightarrow(0,1)\) assigning to each \(x\in\mathbb{R}^{d}\) a probability \(D(x)\) interpreted as the likelihood that \(x\) is real. Two affine layers with a LeakyReLU activation \(\phi(t)=\max(t,\alpha t)\), \(\alpha\in(0,1)\), define an internal representation:
\begin{equation}
u_1=W_{1}^{(D)}x+b_{1}^{(D)},\qquad v_1=\phi(u_1),
\end{equation}
\begin{equation}
u_2=W_{2}^{(D)}v_1+b_{2}^{(D)},\qquad h^{(D)}=\phi(u_2),
\end{equation}
where \(W_{1}^{(D)}\in\mathbb{R}^{k_1\times d}\), \(b_{1}^{(D)}\in\mathbb{R}^{k_1}\), \(W_{2}^{(D)}\in\mathbb{R}^{k_2\times k_1}\), and \(b_{2}^{(D)}\in\mathbb{R}^{k_2}\) are trainable parameters, and \(k_1,k_2\in\mathbb{N}\) are hidden dimensions. During training, dropout with rate \(p\in(0,1)\) is applied to \(h^{(D)}\). Let \(\delta\in\{0,1\}^{k_2}\) be a mask with independent components \(\delta_j\sim\mathrm{Bernoulli}(1-p)\), and define
\begin{equation}
\tilde{h}^{(D)}=\frac{\delta\odot h^{(D)}}{1-p},
\end{equation}
where \(\odot\) denotes element-wise multiplication. The discriminator logit and output are
\begin{equation}
\begin{aligned}
s&=(w^{(D)})^\top\tilde{h}^{(D)}+b^{(D)},\\
D(x)&=\sigma(s)=\frac{1}{1+e^{-s}},
\end{aligned}
\end{equation}
where \(w^{(D)}\in\mathbb{R}^{k_2}\) and \(b^{(D)}\in\mathbb{R}\) are trainable parameters.
\subsection{Adversarial Training with Adaptive Regularization}

Adversarial learning is carried out in the bounded representation space \([-1,1]^d\) produced by preprocessing. Let \(p_{\mathrm{data}}\) denote the empirical distribution of real samples \(x\in\mathbb{R}^{d}\), and let \(p(z)\) denote a fixed latent prior over \(z\in\mathbb{R}^{m}\). At each iteration, a mini-batch \(\{x_i\}_{i=1}^{B}\sim p_{\mathrm{data}}\) and a mini-batch \(\{z_i\}_{i=1}^{B}\sim p(z)\) are drawn, and generated samples are obtained as \(\hat{x}_i=G(z_i)\). The discriminator is optimized to distinguish \(x\) from \(\hat{x}\), while the generator is optimized to reduce the discriminator's ability to separate the two under the same input model.

\subsubsection{Instance noise and projection}
To regularize the discriminator and mitigate early overfitting, instance noise is injected into both real and generated samples before scoring. For any \(u\in\mathbb{R}^{d}\), define the noisy, bounded version
\begin{equation}
\tilde{u}=\Pi_{[-1,1]^d}\!\left(u+\sigma\,\varepsilon\right),\qquad
\varepsilon\sim\mathcal{N}(0,I_d),
\label{eq:inst_noise}
\end{equation}
where \(\sigma\ge 0\) is the noise scale, \(I_d\) is the \(d\times d\) identity matrix, and \(\Pi_{[-1,1]^d}(\cdot)\) denotes element-wise clipping onto \([-1,1]^d\). In particular, \(\tilde{x}\) and \(\tilde{\hat{x}}\) denote the noisy versions of a real sample \(x\) and a generated sample \(\hat{x}=G(z)\), respectively.

\subsubsection{Discriminator objective}
The discriminator output \(D(\cdot)\in(0,1)\) is interpreted as the probability that its input is real. Using binary cross-entropy \(\ell:(0,1)\times\{0,1\}\to\mathbb{R}_{\ge 0}\),
\begin{equation}
\ell(u,y)=-y\log(u)-(1-y)\log(1-u),
\label{eq:bce}
\end{equation}
the discriminator is trained on noisy inputs as
\begin{equation}
\begin{aligned}
\mathcal{L}_D
&=
\mathbb{E}_{x\sim p_{\mathrm{data}}}\!\big[\ell(D(\tilde{x}),1)\big]
+
\mathbb{E}_{z\sim p(z)}\!\big[\ell(D(\tilde{\hat{x}}),0)\big],\\
& \text{with} \quad
\tilde{\hat{x}}=\Pi_{[-1,1]^d}\!\left(G(z)+\sigma\,\varepsilon\right),
\end{aligned}
\label{eq:ld}
\end{equation}
where \(\varepsilon\sim\mathcal{N}(0,I_d)\) is sampled independently of \(z\).

\subsubsection{Generator objective with label smoothing}
The generator is trained to increase discriminator scores on generated samples. To avoid overly sharp targets, the positive label is smoothed to \(\gamma\in(0,1]\), yielding the adversarial generator term
\begin{equation}
\begin{aligned}
\mathcal{L}_{G}^{\mathrm{adv}}
&=
\mathbb{E}_{z\sim p(z)}\!\big[\ell(D(\tilde{\hat{x}}),\gamma)\big],\\
& \text{with} \quad
\tilde{\hat{x}}=\Pi_{[-1,1]^d}\!\left(G(z)+\sigma\,\varepsilon\right),
\end{aligned}
\label{eq:lg_adv}
\end{equation}
with \(\varepsilon\sim\mathcal{N}(0,I_d)\). This uses the same noisy input model as the discriminator loss in Eq.~\eqref{eq:ld}.

\subsubsection{Feature matching regularization}
Let \(f:\mathbb{R}^{d}\to\mathbb{R}^{k}\) denote the discriminator feature map extracted from a fixed intermediate layer, with \(k\in\mathbb{N}\). Feature matching penalizes the \(\ell_1\) discrepancy between mean discriminator features of real and generated samples:
\begin{equation}
\mathcal{L}_{G}^{\mathrm{FM}}
=
\lambda_{\mathrm{FM}}
\left\|
\mathbb{E}_{x\sim p_{\mathrm{data}}}\!\big[f(x)\big]
-
\mathbb{E}_{z\sim p(z)}\!\big[f(G(z))\big]
\right\|_{1},
\label{eq:lg_fm}
\end{equation}
where \(\lambda_{\mathrm{FM}}\ge 0\) is a weighting coefficient. Features are computed on non-noisy samples to match the underlying distributional representations rather than their perturbed versions. Here, \(f(x)\) denotes the discriminator activations after the second LeakyReLU layer and before dropout, with dropout disabled during feature extraction to ensure a deterministic matching signal.

\subsubsection{Moment matching regularization}
To align low-order statistics directly in representation space, let \(\mu_B(\cdot)\in\mathbb{R}^{d}\) and \(s_B(\cdot)\in\mathbb{R}^{d}\) denote the component-wise mean and standard deviation computed over a mini-batch of size \(B\). For a batch \(\{u_i\}_{i=1}^{B}\), define
\begin{equation}\label{eq:batch_stats}
\begin{split}
\mu_B(u)&=\frac{1}{B}\sum_{i=1}^{B}u_i,\\
\raisetag{11ex}
s_B(u)&=\sqrt{\frac{1}{B}\sum_{i=1}^{B}\bigl(u_i-\mu_B(u)\bigr)\odot\bigl(u_i-\mu_B(u)\bigr)+\varepsilon_{\sigma}\mathbf{1}},
\end{split}
\end{equation}
where \(\odot\) denotes element-wise multiplication, \(\mathbf{1}\in\mathbb{R}^{d}\) is the all-ones vector, and \(\varepsilon_{\sigma}>0\) is a small constant. The moment-matching term is then
\begin{equation}
\mathcal{L}_{G}^{\mathrm{MM}}
=
\alpha\,
\big\|\mu_B(x)-\mu_B(\hat{x})\big\|_{1}
+
\beta\,
\big\|s_B(x)-s_B(\hat{x})\big\|_{1},
\label{eq:lg_mm}
\end{equation}
where \(x=\{x_i\}_{i=1}^{B}\) is a real mini-batch, \(\hat{x}=\{\hat{x}_i\}_{i=1}^{B}\) is a generated mini-batch, and \(\alpha,\beta\ge 0\) are weighting coefficients.

\subsubsection{Total generator loss and constrained updates}
The overall generator objective combines adversarial learning with the two regularizers:
\begin{equation}
\mathcal{L}_G
=
\mathcal{L}_{G}^{\mathrm{adv}}
+
\mathcal{L}_{G}^{\mathrm{FM}}
+
\mathcal{L}_{G}^{\mathrm{MM}}.
\label{eq:lg_total}
\end{equation}
Let \(\theta_D\) and \(\theta_G\) denote discriminator and generator parameters. Training alternates between a discriminator step minimizing \(\mathcal{L}_D(\theta_D;\theta_G)\) in Eq.~\eqref{eq:ld} and a generator step minimizing \(\mathcal{L}_G(\theta_G;\theta_D)\) in Eq.~\eqref{eq:lg_total}. To stabilize generator optimization, gradient norm clipping is applied:
\begin{equation}
\left\|\nabla_{\theta_G}\mathcal{L}_G\right\|_2 \le c,
\label{eq:clip}
\end{equation}
where \(c>0\) is the clipping threshold.

\subsubsection{Adaptive adjustment of regularization intensity}
Regularization controls are adapted at evaluation checkpoints to maintain informative gradients and avoid an over-confident discriminator. Let \(\sigma\) denote the instance-noise scale in Eq.~\eqref{eq:inst_noise}, let \(\gamma\) denote the label-smoothing target in Eq.~\eqref{eq:lg_adv}, and let \(p\in(0,1)\) denote the discriminator dropout rate defined in the previous subsection. At each checkpoint, summary indicators derived from discriminator behavior and real-vs-generated sample statistics are used to update \((\sigma,\gamma,p)\) within predefined bounds, increasing regularization when discrimination becomes too easy and relaxing it when the discriminator signal becomes uninformative.
\subsection{Evaluation Protocol and Distributional Similarity Metrics}

Evaluation compares real and generated samples in the learned representation space using two complementary criteria: marginal distributional similarity computed independently per feature dimension and external detectability measured by a post hoc classifier. Both criteria are computed from real samples drawn from the empirical distribution and synthetic samples obtained by sampling the latent prior and applying the trained generator, ensuring consistency with the end-to-end generation mechanism used during training.

Let $x\in\mathbb{R}^{d}$ denote a real sample drawn from $p_{\mathrm{data}}$, and let $z\in\mathbb{R}^{m}$ denote a latent sample drawn from a fixed prior $p(z)$. A generated sample is $\hat{x}=G(z)\in\mathbb{R}^{d}$. For each evaluation round, we select $n\in\mathbb{N}$ real samples $\{x_i\}_{i=1}^{n}$ and draw $n$ latent samples $\{z_i\}_{i=1}^{n}$ to form generated samples $\{\hat{x}_i\}_{i=1}^{n}$ with $\hat{x}_i=G(z_i)$. When instance noise is used during evaluation, we apply the same bounded perturbation operator used in training. Specifically, with noise scale $\sigma\ge 0$ and i.i.d.\ Gaussian vectors $\varepsilon_i,\varepsilon_i'\sim\mathcal{N}(0,I_d)$, we define
\begin{equation}
\begin{aligned}
\tilde{x}_i&=\Pi_{[-1,1]^d}\!\left(x_i+\sigma\,\varepsilon_i\right),\\
\tilde{\hat{x}}_i&=\Pi_{[-1,1]^d}\!\left(\hat{x}_i+\sigma\,\varepsilon_i'\right),
\end{aligned}
\end{equation}
where $I_d$ is the $d\times d$ identity matrix and $\Pi_{[-1,1]^d}$ denotes element-wise projection onto $[-1,1]^d$. Unless stated otherwise, we set $\sigma$ at evaluation equal to the instance-noise value used during training at the corresponding checkpoint, so that evaluation is performed under the same perturbation model.

\subsubsection{External detectability via AUC}
To quantify real-vs-synthetic detectability, we compute the AUC-ROC of an independently trained classifier that distinguishes real fraud samples from synthetic fraud samples generated by a \emph{frozen} generator. After Q-SYNTH (or a baseline generator) finishes training, we freeze its parameters and sample $n$ synthetic fraud records $\{\hat{x}_i\}_{i=1}^{n}$ by drawing latent variables $\{z_i\}_{i=1}^{n}\sim p(z)$ and applying $\hat{x}_i = G(z_i)$. We also draw $n$ real fraud samples $\{x_i\}_{i=1}^{n}$ from a held-out fraud split.

We then train a logistic regression detector $C:\mathbb{R}^d\rightarrow(0,1)$ on a training partition of the combined set
$\{(x_i,1)\}_{i=1}^{n}\cup\{(\hat{x}_i,0)\}_{i=1}^{n}$,
and evaluate it on a disjoint test partition. Let $s_i=C(x_i)$ and $\hat{s}_i=C(\hat{x}_i)$ denote the detector scores on real and synthetic fraud test samples, respectively. We form the label vector $y\in\{0,1\}^{2n}$ and score vector $\pi\in(0,1)^{2n}$ as
\begin{equation}
\begin{aligned}
y&=(\underbrace{1,\dots,1}_{n\ \text{times}},\underbrace{0,\dots,0}_{n\ \text{times}}),\\
\pi&=(s_1,\dots,s_n,\hat{s}_1,\dots,\hat{s}_n).
\end{aligned}
\end{equation}

The AUC-ROC, denoted $\mathrm{AUC}\in[0,1]$, equals the probability that a randomly selected real fraud sample receives a higher detector score than a randomly selected synthetic fraud sample. Values close to $0.5$ indicate low detection, while values farther from $0.5$ indicate greater detectability. We report both $\mathrm{AUC}$ and the detectability gap $|\mathrm{AUC}-0.5|$.

\subsubsection{Per-dimension Kolmogorov--Smirnov statistics}
Marginal similarity is assessed using the two-sample Kolmogorov--Smirnov statistic computed per coordinate. For each $j\in\{1,\dots,d\}$, let $\{x_{i,j}\}_{i=1}^{n}$ and $\{\hat{x}_{i,j}\}_{i=1}^{n}$ be the real and generated samples restricted to dimension $j$, and let $F_{n,j}$ and $\hat{F}_{n,j}$ be the corresponding empirical CDFs. The KS statistic is
\begin{equation}
K_j=\sup_{t\in\mathbb{R}} \left|F_{n,j}(t)-\hat{F}_{n,j}(t)\right|,
\end{equation}
and the associated KS p-value is denoted $P_j\in(0,1]$. We report robust aggregates across dimensions using medians
\begin{equation}
\begin{aligned}
K_{\mathrm{med}}&=\mathrm{median}\left(\{K_j\}_{j=1}^{d}\right),\\
P_{\mathrm{med}}&=\mathrm{median}\left(\{P_j\}_{j=1}^{d}\right).
\end{aligned}
\end{equation}

\subsubsection{Per-dimension Wasserstein distances}
We additionally compute the 1-Wasserstein distance per coordinate. For each $j\in\{1,\dots,d\}$, let $W_j\ge 0$ denote the empirical 1-Wasserstein distance between $\{x_{i,j}\}_{i=1}^{n}$ and $\{\hat{x}_{i,j}\}_{i=1}^{n}$. To summarize across dimensions, we report the median and an upper-tail empirical quantile. For a chosen percentile level $q\in(0,100)$ (e.g., $q=75$), define
\begin{equation}
\begin{aligned}
W_{\mathrm{med}}&=\mathrm{median}\left(\{W_j\}_{j=1}^{d}\right),\\
W_{q}&=Q_{q}\left(\{W_j\}_{j=1}^{d}\right),
\end{aligned}
\end{equation}
where $Q_q(\cdot)$ denotes the empirical $q$-th percentile. The median captures typical marginal discrepancy, while $W_q$ summarizes larger mismatches present in a subset of coordinates.
\section{Results and Discussion}\label{sec4}

\subsection{Experimental Setup}
\label{subsec:exp_setup}

All experiments are conducted under a single, fixed protocol to ensure that performance differences reflect the augmentation strategy and the downstream learner, rather than variations in preprocessing, model capacity, or training schedule. Unless stated otherwise, each method uses its standard default settings, with only the hyperparameters in Tables~\ref{tab:exp_setup} and~\ref{tab:hparams_methods_clean_lines} explicitly set or tuned. Table~\ref{tab:exp_setup} summarizes the full Q-SYNTH configuration, covering preprocessing constants, quantum generator design, discriminator architecture, optimization schedule, regularization terms, and evaluation cadence.

To make the adversarial baselines as comparable as possible, the classical GAN and Q-SYNTH are matched in latent/representation dimension and constrained to similar generator capacity by aligning the number of trainable parameters as closely as feasible, given the architectural mismatch between neural generators and parameterized quantum circuits. Additional augmentation methods are included in the motivation summary (Fig.~\ref{histogram}) under the same preprocessing, data splits, and evaluation protocol to provide broader context. However, the main quantitative discussion focuses on SMOTE and the classical GAN as primary reference baselines to keep comparisons controlled and interpretable: SMOTE provides a strong non-adversarial oversampling method that preserves local feature statistics via interpolation, while the classical GAN is the closest adversarial counterpart and isolates the effect of replacing a classical generator with a quantum generator within the same generation-discrimination framework.

Downstream performance is evaluated primarily with a QNN classifier to reflect the quantum-centric goal of the study, and is complemented with representative classical classifiers (ANN, logistic regression, random forest, and XGBoost) to verify that observed trends are not specific to a single learner. The QNN employs angle embedding and a StronglyEntanglingLayers\footnote{https://docs.pennylane.ai/en/stable/code/api/pennylane.StronglyEntanglingLayers.html} ansatz followed by a classical decision head, and the key downstream hyperparameters are listed in Table~\ref{tab:hparams_methods_clean_lines}. All classifiers are trained and assessed under identical data splits and metric definitions to enable consistent comparisons across augmentation strategies.

\begin{table}[hbpt]
\centering
\caption{Experimental configuration of Q-SYNTH pipeline.}
\label{tab:exp_setup}
\small
\resizebox{\linewidth}{!}{%
\begin{tabular}{ll}
\toprule
\rowcolor{gray!18}
\textbf{Component / hyperparameter} & \textbf{Setting} \\
\midrule
Normalization floor $\varepsilon$ & $10^{-8}$ \\
Quantum backend & Pennylane (default.qubit) \cite{bergholm2018pennylane} \\
Quantum circuit layers & $8$ \\
Generator MLP hidden size & $32$ \\
Discriminator hidden sizes & $16 \rightarrow 8$ \\
Discriminator activation & LeakyReLU $(\alpha=0.2)$ \\
Initial discriminator dropout $p$ & $0.10$ \\
Optimizer & Adam $(\beta_1=0.5,\ \beta_2=0.9)$ \\
Learning rates (G / D) & $7\times 10^{-4}$ / $2\times 10^{-4}$ \\
Batch size & $64$ \\
Training epochs & $100$ \\
Initial generator target label $\gamma$ & $0.88$ \\
Instance noise (base) & $0.014$ \\
Instance noise (end bonus) & $0.016$ \\
Initial adaptive noise increment $\Delta$ & $0.002$ \\
Evaluation frequency & every $10$ epochs \\
Scheduler bounds for $\gamma$ & $[0.80,\ 0.94]$ \\
Scheduler bounds for dropout $p$ & $[0.10,\ 0.16]$ \\
Scheduler step (noise) & $0.004$ \\
Scheduler step ($\gamma$) & $0.02$ \\
Scheduler step (dropout) & $0.03$ \\
Feature-matching weight & $0.10$ \\
Moment-matching weights $(\alpha,\beta)$ & $(0.05,\ 0.03)$ \\
Moment std floor $\varepsilon_{\sigma}$ & $10^{-6}$ \\
Generator gradient clipping $\|\cdot\|_2$ & $1.0$ \\
Evaluation sample size $n$ & $2000$ \\
\bottomrule
\end{tabular}}
\end{table}

\begin{table}[htpbt]
\centering
\caption{Key hyperparameters for each downstream classifier on the balanced dataset (984 fraud + 984 non-fraud). Unless specified, default settings are used.}
\label{tab:hparams_methods_clean_lines}
\small
\resizebox{\linewidth}{!}{%
\begin{tabular}{ll}
\toprule
\rowcolor{gray!18}
\textbf{Model} & \textbf{Key hyperparameters} \\
\midrule
\textbf{QNN} &
Number of qubits: $6$ \\
& Quantum layers: $3$  \\
& Classical layers: $2$ \\
& Optimizer: Adam \\
& Learning rate: $10^{-3}$ \\
& Epochs: $30$ \\
& Batch size: $64$ \\
\midrule
\textbf{ANN} &
Hidden layers: $3$ \\
& Optimizer: Adam \\
& Learning rate: $10^{-3}$ \\
& Epochs: $30$ \\
& Batch size: $64$ \\
\midrule
\textbf{Logistic Regression} &
Maximum iterations: $5000$ \\
\midrule
\textbf{Random Forest} &
Number of estimators: $200$ \\
\midrule
\textbf{XGBoost} &
Number of estimators: $300$ \\
& Maximum depth: $5$ \\
& Learning rate: $0.1$ \\
& Subsample ratio: $0.8$ \\
& Column subsample ratio per tree: $0.8$ \\
\bottomrule
\end{tabular}}
\end{table}

\subsection{Training Dynamics and Convergence}

\begin{figure}[htpb]
    \centering
    \includegraphics[width=1\linewidth]{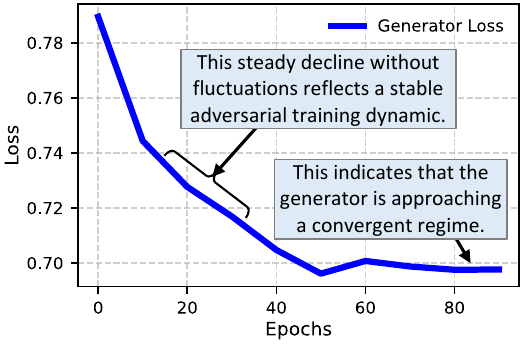}
    \caption{Generator loss over training epochs.}
    \label{fig:gen_loss}
\end{figure}
We examine the training dynamics to assess whether adversarial optimization remains qualitatively stable and whether the generator progressively improves its approximation of the target fraud distribution in the learned representation space. The generator loss trend, presented in Fig.~\ref{fig:gen_loss}, starts at approximately (0.789) at early epochs and decreases to about (0.697) by late training (epoch 90). The overall downward trajectory, with minor fluctuations, is consistent with steady progress in generator updates over the reported run.
\begin{figure}[htpb]
    \centering
    \includegraphics[width=1\linewidth]{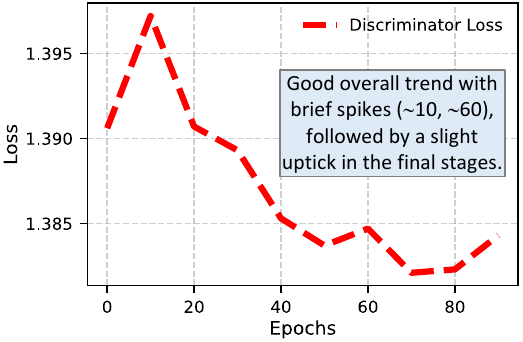}
    \caption{Discriminator loss over training epochs. }
    \label{fig:disc_loss}
\end{figure}

The discriminator loss curve in Fig.~\ref{fig:disc_loss} remains relatively high and nearly stationary throughout training. Starting around (1.390), it exhibits small oscillations but no sustained drift, suggesting that the discriminator reaches a relatively steady regime. In contrast to the generator’s gradual decrease, this near-constant behavior is qualitatively consistent with adversarial training in which neither player visibly collapses or dominates.

Overall, the loss curves suggest a controlled training process in the reported run: the generator improves gradually, while the discriminator remains relatively stable. These plots should be interpreted as qualitative indicators of training behavior rather than as a complete characterization of optimization robustness.

\subsection{Fidelity of Generated Fraud Samples}
\label{subsec:fidelity}

\begin{table*}[htpbt]
\centering
\caption{Distributional similarity between real fraud samples and synthetic fraud samples produced by different augmentation techniques. Lower is better for KS and Wasserstein distances; higher is better for KS $p$-value; AUC (closer to 0.5 is better).}
\label{statis}

\resizebox{\linewidth}{!}{%
\begin{tabular}{lccccc}
\toprule
\rowcolor{headgray}
\textbf{Technique} &
\textbf{KS Median $\downarrow$} &
\textbf{KS $p$-value $\uparrow$} &
\textbf{Wass. Median $\downarrow$} &
\textbf{Wass. P75 $\downarrow$} &
\textbf{AUC} \\
\midrule
SMOTE & 0.034553 & 0.931076 & 0.156992 & 0.250870 & 0.518764 \\
\rowcolor{rowgray}
GAN   & 0.184959 & $1.507359\times 10^{-7}$ & 1.514651 & 2.24196 & 0.51155 \\
\rowcolor{qganrow}
\textbf{Q-SYNTH} & 0.069000 & 0.196000 & 0.498000 & 1.00200 & 0.47500 \\
\bottomrule
\end{tabular}%
}
\end{table*}

This experiment evaluates whether Q-SYNTH improves the fidelity of adversarially generated fraud samples relative to a classical GAN baseline, while also contrasting with SMOTE as a widely used non-adversarial oversampling reference. As summarized in Table~\ref{statis}, SMOTE achieves the closest marginal agreement with real fraud samples (lowest KS and Wasserstein distances, highest KS $p$-value), which is expected because interpolation-based oversampling tends to preserve local feature-wise statistics. In contrast, the classical GAN exhibits substantially larger marginal discrepancies (KS median 0.184959 and Wasserstein median 1.514651), indicating a weaker match to the empirical fraud marginals under the same protocol. Q-SYNTH reduces these discrepancies relative to the classical GAN across all reported marginal indicators (KS median 0.069000; Wasserstein median 0.498000; P75 1.00200), indicating a narrower marginal distribution gap under the reported protocol.

\begin{figure}[htpb]
    \centering
    \includegraphics[width=1\linewidth]{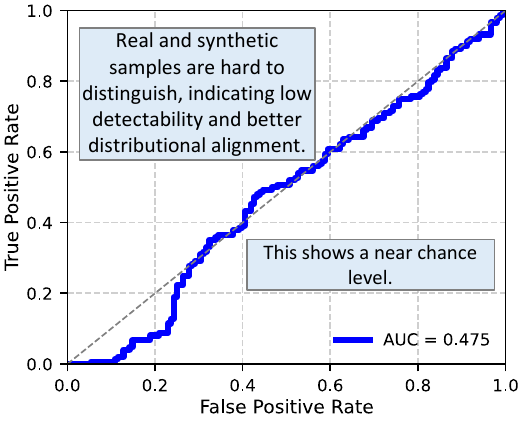}
\caption{ROC curve of an external logistic regression detector trained after generator training is complete to distinguish held-out real fraud from Q-SYNTH-generated fraud, reported as AUC-ROC ($\mathrm{AUC}=0.475$). Values closer to $0.5$ indicate reduced detectability.}
    \label{auc}
\end{figure}

In addition to marginal similarity, Table~\ref{statis} reports real-vs-synthetic detectability using AUC-ROC from an \emph{external} logistic regression detector trained after generator training is complete. Specifically, for each augmentation method, we freeze the trained generator, generate synthetic fraud samples, and train logistic regression to distinguish held-out real fraud from synthetic fraud; AUC is computed on a disjoint test split. Deviations from $0.5$ indicate increased detectability, while values near $0.5$ indicate low detectability under this independent detector. SMOTE and GAN remain close to low detectability (AUC $0.518764$ and $0.51155$). Q-SYNTH yields $\mathrm{AUC}=0.475$, with detectability gap $|\mathrm{AUC}-0.5|=0.025$, indicating near-chance detectability. The corresponding ROC curve is shown in Fig.~\ref{auc}. Overall, the fidelity results highlight a structured trade-off: SMOTE best preserves feature-wise marginals, while Q-SYNTH improves substantially over the classical GAN in marginal fidelity while maintaining comparable detectability under an independent classifier.

\subsection{Downstream Fraud Detection Performance}
\label{subsec:downstream}

Downstream evaluation measures how augmentation changes fraud-class detection when training classifiers on the augmented data. Results are reported for a QNN (the primary setting) and for representative classical models to verify that observed trends are not specific to a single learner.

\begin{table*}[htpb]
\centering
\caption{Downstream fraud detection performance across classifiers under different augmentation strategies. Metrics are reported for the non-fraud class ($y=0$) and the fraud class ($y=1$), with AUC and accuracy. Q-SYNTH rows are highlighted for readability.}
\label{tab:downstream_all_models}

\resizebox{\textwidth}{!}{%
\begin{tabular}{llcccccccc}
\toprule
\rowcolor{headgray}
\textbf{Model} & \textbf{Augmentation} &
\multicolumn{3}{c}{\textbf{$y=0$ (Non-fraud)}} &
\multicolumn{3}{c}{\textbf{$y=1$ (Fraud)}} &
\textbf{AUC} & \textbf{Acc.} \\
\rowcolor{headgray}
& &
\textbf{Prec.} & \textbf{Rec.} & \textbf{F1} &
\textbf{Prec.} & \textbf{Rec.} & \textbf{F1} &
& \\
\midrule

\multirow{5}{*}{\textbf{QNN}}
& Imbalanced & 0.9995 & 0.9998 & 0.9997 & 0.8699 & 0.7230 & 0.7897 & 0.9582 & 0.9993 \\
\rowcolor{rowgray}
& Balanced   & 0.8810 & 1.0000 & 0.9367 & 1.0000 & 0.8649 & 0.9275 & 0.9838 & 0.9324 \\
& SMOTE      & 0.8963 & 0.9932 & 0.9423 & 0.9924 & 0.8847 & 0.9355 & 0.9778 & 0.9391 \\
\rowcolor{rowgray}
& GAN        & 0.9363 & 0.9932 & 0.9639 & 0.9928 & 0.9322 & 0.9615 & 0.9904 & 0.9628 \\
\rowcolor{qganrow}
& \textbf{Q-SYNTH} & 0.9216 & 0.9932 & 0.9561 & 0.9926 & \textbf{0.9153} & \textbf{0.9524} & 0.9896 & 0.9543 \\
\midrule

\multirow{5}{*}{\textbf{ANN}}
& Imbalanced & 0.9996 & 0.9997 & 0.9997 & 0.8248 & 0.7635 & 0.7930 & 0.9522 & 0.9993 \\
\rowcolor{rowgray}
& Balanced   & 0.9063 & 0.9797 & 0.9416 & 0.9779 & 0.8986 & 0.9366 & 0.9842 & 0.9392 \\
& SMOTE      & 0.8988 & 0.9899 & 0.9421 & 0.9887 & 0.8881 & 0.9357 & 0.9781 & 0.9391 \\
\rowcolor{rowgray}
& GAN        & 0.9423 & 0.9932 & 0.9671 & 0.9928 & 0.9390 & 0.9652 & 0.9881 & 0.9662 \\
\rowcolor{qganrow}
& \textbf{Q-SYNTH} & 0.9448 & 0.9831 & 0.9636 & 0.9823 & \textbf{0.9424} & \textbf{0.9619} & 0.9885 & 0.9628 \\
\midrule

\multirow{5}{*}{\textbf{LogReg}}
& Imbalanced & 0.9993 & 0.9998 & 0.9996 & 0.8558 & 0.6014 & 0.7063 & 0.9583 & 0.9991 \\
\rowcolor{rowgray}
& Balanced   & 0.9231 & 0.9730 & 0.9474 & 0.9714 & 0.9189 & 0.9444 & 0.9839 & 0.9459 \\
& SMOTE      & 0.9140 & 0.9696 & 0.9410 & 0.9675 & 0.9085 & 0.9371 & 0.9829 & 0.9391 \\
\rowcolor{rowgray}
& GAN        & 0.9510 & 0.9831 & 0.9668 & 0.9825 & 0.9492 & 0.9655 & 0.9912 & 0.9662 \\
\rowcolor{qganrow}
& \textbf{Q-SYNTH} & 0.9474 & 0.9730 & 0.9600 & 0.9721 & \textbf{0.9458} & \textbf{0.9588} & 0.9906 & 0.9594 \\
\midrule

\multirow{5}{*}{\textbf{RF}}
& Imbalanced & 0.9996 & 1.0000 & 0.9998 & 0.9669 & 0.7905 & 0.8699 & 0.9366 & 0.9996 \\
\rowcolor{rowgray}
& Balanced   & 0.9276 & 0.9527 & 0.9400 & 0.9514 & 0.9257 & 0.9384 & 0.9852 & 0.9392 \\
& SMOTE      & 0.9465 & 0.9561 & 0.9513 & 0.9555 & 0.9458 & 0.9506 & 0.9885 & 0.9509 \\
\rowcolor{rowgray}
& GAN        & 0.9510 & 0.9831 & 0.9668 & 0.9825 & 0.9492 & 0.9655 & 0.9912 & 0.9662 \\
\rowcolor{qganrow}
& \textbf{Q-SYNTH} & 0.9507 & 0.9764 & 0.9633 & 0.9756 & \textbf{0.9492} & \textbf{0.9622} & 0.9912 & 0.9628 \\
\midrule

\multirow{5}{*}{\textbf{XGB}}
& Imbalanced & 0.9996 & 0.9998 & 0.9997 & 0.8702 & 0.7703 & 0.8172 & 0.9648 & 0.9994 \\
\rowcolor{rowgray}
& Balanced   & 0.9272 & 0.9459 & 0.9365 & 0.9448 & 0.9257 & 0.9352 & 0.9841 & 0.9358 \\
& SMOTE      & 0.9465 & 0.9561 & 0.9513 & 0.9555 & 0.9458 & 0.9506 & 0.9890 & 0.9509 \\
\rowcolor{rowgray}
& GAN        & 0.9538 & 0.9764 & 0.9649 & 0.9757 & 0.9525 & 0.9640 & 0.9896 & 0.9645 \\
\rowcolor{qganrow}
& \textbf{Q-SYNTH} & 0.9564 & 0.9628 & 0.9596 & 0.9625 & \textbf{0.9559} & \textbf{0.9592} & 0.9893 & 0.9594 \\
\bottomrule
\end{tabular}%
}
\end{table*}

Across models, augmentation primarily affects the minority class, while majority-class performance remains near-saturated. Relative to the balanced baseline and to SMOTE, Q-SYNTH generally improves fraud recall and fraud F1-score across the reported models, suggesting that Q-SYNTH-generated samples add minority-class structure beyond local interpolation. At the same time, the classical GAN attains the strongest downstream scores in several configurations. Q-SYNTH nevertheless remains competitive and often close in fraud recall and F1-score, indicating that the hybrid generator produces samples that are useful for classification even when it does not maximize downstream performance. When considered together with the fidelity analysis in Table~\ref{statis}, Q-SYNTH offers a favorable trade-off: it improves markedly over the classical GAN in marginal similarity while retaining competitive downstream performance across both quantum and classical classifiers.

\subsection{Scaling Analysis}
\label{subsec:scaling}
\begin{table*}[htpbt]
\centering
\caption{Scaling analysis for a QNN trained with Q-SYNTH-generated fraud samples. Performance is reported under balanced and imbalanced evaluation settings. Blue shading emphasizes fraud-class Recall and F1.}
\label{tab:scaling_qnn_readable}

\resizebox{\textwidth}{!}{%
\begin{tabular}{l l c c c c c c c c}
\toprule
\rowcolor{headgray}
 & \textbf{Case (Synthetic Fraud)} &
\multicolumn{3}{c}{\textbf{$y=0$ (Non-fraud)}} &
\multicolumn{3}{c}{\textbf{$y=1$ (Fraud)}} &
\textbf{AUC} & \textbf{Acc.} \\
\rowcolor{headgray}
& &
\textbf{Prec.} & \textbf{Rec.} & \textbf{F1} &
\textbf{Prec.} & \textbf{Rec.} & \textbf{F1} &
& \\
\midrule

\multirow{6}{*}{\rotatebox{90}{\textbf{Balanced}}}
& Real Only (0)
& 0.8810 & 1.0000 & 0.9367
& 1.0000 & \hcell{0.8649} & \hcell{0.9275}
& 0.9838 & 0.9324 \\

\grayrow 10\% Synthetic (49)
& \cellcolor{rowgray}0.8571 & \cellcolor{rowgray}0.9939 & \cellcolor{rowgray}0.9205
& \cellcolor{rowgray}0.9926 & \cellcolor{rowgray}\hcell{0.8333} & \cellcolor{rowgray}\hcell{0.9060}
& \cellcolor{rowgray}0.9733 & \cellcolor{rowgray}0.9138 \\

& 25\% Synthetic (123)
& 0.8966 & 0.9838 & 0.9381
& 0.9819 & \hcell{0.8859} & \hcell{0.9314}
& 0.9845 & 0.9350 \\

\grayrow 50\% Synthetic (246)
& \cellcolor{rowgray}0.9283 & \cellcolor{rowgray}0.9910 & \cellcolor{rowgray}0.9586
& \cellcolor{rowgray}0.9903 & \cellcolor{rowgray}\hb{0.9231} & \cellcolor{rowgray}\hb{0.9555}
& \cellcolor{rowgray}0.9881 & \cellcolor{rowgray}0.9571 \\

& 100\% Synthetic (492)
& 0.9216 & 0.9932 & 0.9561
& 0.9926 & \hcell{0.9153} & \hcell{0.9524}
& 0.9896 & 0.9543 \\

\grayrow 100\% Synthetic Only (492)
& \cellcolor{rowgray}0.9308 & \cellcolor{rowgray}1.0000 & \cellcolor{rowgray}0.9642
& \cellcolor{rowgray}1.0000 & \cellcolor{rowgray}\hcell{0.9257} & \cellcolor{rowgray}\hb{0.9614}
& \cellcolor{rowgray}0.9991 & \cellcolor{rowgray}0.9628 \\

\midrule

\multirow{6}{*}{\rotatebox{90}{\textbf{Imbalanced}}}
& Real Only (0)
& 0.9284 & 1.0000 & 0.9629
& 1.0000 & \hcell{0.7500} & \hcell{0.8571}
& 0.9855 & 0.9411 \\

\grayrow 10\% Synthetic (49)
& \cellcolor{rowgray}0.9322 & \cellcolor{rowgray}1.0000 & \cellcolor{rowgray}0.9649
& \cellcolor{rowgray}1.0000 & \cellcolor{rowgray}\hcell{0.7840} & \cellcolor{rowgray}\hcell{0.8789}
& \cellcolor{rowgray}0.9818 & \cellcolor{rowgray}0.9456 \\

& 25\% Synthetic (123)
& 0.9194 & 0.9979 & 0.9570
& 0.9931 & \hcell{0.7730} & \hcell{0.8693}
& 0.9773 & 0.9353 \\

\grayrow 50\% Synthetic (246)
& \cellcolor{rowgray}0.9483 & \cellcolor{rowgray}0.9938 & \cellcolor{rowgray}0.9705
& \cellcolor{rowgray}0.9849 & \cellcolor{rowgray}\hcell{0.8829} & \cellcolor{rowgray}\hcell{0.9311}
& \cellcolor{rowgray}0.9880 & \cellcolor{rowgray}0.9587 \\

& 100\% Synthetic (492)
& 0.9389 & 0.9917 & 0.9645
& 0.9851 & \hb{0.8953} & \hb{0.9381}
& 0.9859 & 0.9549 \\

\grayrow 100\% Synthetic Only (492)
& \cellcolor{rowgray}0.9658 & \cellcolor{rowgray}1.0000 & \cellcolor{rowgray}0.9826
& \cellcolor{rowgray}1.0000 & \cellcolor{rowgray}\hcell{0.8851} & \cellcolor{rowgray}\hcell{0.9391}
& \cellcolor{rowgray}0.9995 & \cellcolor{rowgray}0.9729 \\

\bottomrule
\end{tabular}%
}
\end{table*}

This study examines how the fraction of Q-SYNTH-generated fraud samples affects downstream detection during QNN training. The synthetic-to-real fraud ratio is varied by injecting \(10\%\), \(25\%\), \(50\%\), and \(100\%\) synthetic fraud samples into the training set, and a synthetic-only condition is additionally reported for the fraud class. Results are evaluated under two settings: a balanced protocol and an imbalanced protocol, with metrics reported separately for the non-fraud class (\(y=0\)) and the fraud class (\(y=1\)).

As reported in Table~\ref{tab:scaling_qnn_readable}, the balanced setting shows non-monotonic behavior at small injection ratios: introducing \(10\%\) synthetic fraud samples reduces performance relative to the real-only baseline for both classes, indicating that a small synthetic fraction may not provide a sufficiently consistent minority-class signal. At \(25\%\), performance recovers to approximately the real-only level. Increasing the fraction to \(50\%\) yields the strongest balanced performance across most indicators, notably fraud recall and fraud F1-score, while \(100\%\) maintains similarly high results. The synthetic-only condition also remains strong within the balanced regime, suggesting that Q-SYNTH-generated fraud samples preserve task-relevant structure under the present protocol, though this should not be interpreted as evidence that real fraud data can be broadly replaced in operational settings.

Under the imbalanced setting, non-fraud performance (\(y=0\)) remains near-saturated across all cases, whereas fraud-class detection improves as the synthetic fraction increases. In particular, fraud recall and fraud F1-score rise substantially at (50\%) and above, consistent with the generator providing broader coverage of minority-class patterns that are otherwise underrepresented. The synthetic-only condition also achieves strong fraud metrics while keeping the majority-class performance high, and it produces the highest AUC values in this experiment. Overall, the scaling trends indicate that moderate-to-high injection ratios (\(\ge 25\%\), and especially around \(50\%\)) provide the most reliable gains for fraud detection, while very small synthetic fractions may be insufficient to shift the learned decision boundary.

These results support the use of Q-SYNTH as a controllable augmentation mechanism for strengthening minority-class learning in the QNN within the present preprocessing and evaluation setup.

\subsection{Discussion}

The reported results support the central objective of Q-SYNTH: improving the quality of adversarial fraud augmentation while retaining competitive downstream performance under class imbalance. The fidelity analysis shows that Q-SYNTH substantially reduces the marginal distribution gap relative to a classical GAN, as reflected in lower KS and Wasserstein distances across features, while remaining low on real-vs-synthetic detectability. This combination suggests that the hybrid generator mitigates some of the marginal artifacts that can arise in classical adversarial training on tabular fraud data, improving feature-wise fidelity without making synthetic samples trivially distinguishable by an independent detector.

Downstream evaluation further indicates that Q-SYNTH-generated fraud samples are beneficial for minority-class detection across both the QNN and representative classical learners. Relative to the balanced baseline and to SMOTE, Q-SYNTH generally improves fraud recall and fraud F1-score across the reported models, suggesting that adversarially generated samples enrich decision-relevant structure beyond local interpolation. At the same time, the classical GAN attains the strongest downstream scores in several configurations. Q-SYNTH nevertheless remains competitive and often close in fraud recall and F1-score, while simultaneously offering a clear advantage in marginal fidelity. This highlights the central empirical trade-off in the study: classical GANs may maximize downstream performance in some settings, whereas Q-SYNTH provides a more balanced profile by improving distributional agreement while maintaining competitive detection performance.

The scaling study supports this interpretation by showing that Q-SYNTH acts as a controllable augmentation mechanism: performance improvements are most reliable at moderate-to-high injection ratios, typically around (50\%), whereas very small synthetic fractions may be insufficient to shift the decision boundary. The synthetic-only stress test also yields strong performance within the present protocol, indicating that generated fraud samples retain task-relevant signals against real non-fraud instances under the chosen preprocessing and evaluation setup.

Several considerations define the scope of the present study and motivate future work. First, as in other adversarial augmentation frameworks, training behavior depends on hyperparameter choices and initialization, and hybrid QGANs may additionally reflect circuit-optimization and measurement-related variability. Second, the current evaluation focuses on feature-wise distributional fidelity; future extensions with dependence-sensitive metrics would further strengthen the assessment of global sample structure. Third, the present results are obtained within a compressed representation pipeline designed to make hybrid quantum generation tractable for tabular fraud synthesis. Finally, hardware-aware evaluation remains an important next step for understanding performance under realistic NISQ constraints, including noise, limited connectivity, circuit depth, and shot cost. Within this scope, the reported results support Q-SYNTH as a competitive hybrid augmentation approach for imbalanced fraud detection.
\section{Conclusion}\label{sec5}
This paper introduced \textbf{Q-SYNTH}, a hybrid quantum generative adversarial framework for tabular fraud augmentation, in which a parameterized quantum circuit serves as the generator and a classical discriminator drives adversarial learning. Q-SYNTH is designed for minority-class synthesis under severe class imbalance, with the goal of improving fraud detection beyond interpolation-based oversampling alone. Under the reported protocol, Q-SYNTH improves feature-wise distributional agreement relative to a classical GAN baseline while maintaining low real-vs-synthetic detectability. In downstream evaluation, Q-SYNTH-based augmentation generally improves fraud-class recall and F1-score relative to balanced and SMOTE baselines across a QNN and representative classical classifiers, while remaining competitive with a classical GAN in downstream performance. The scaling analysis further indicates that the synthetic injection ratio acts as a practical control variable, with the most reliable gains observed at moderate-to-high augmentation levels. Taken together, these results position Q-SYNTH as a promising hybrid augmentation strategy for imbalanced tabular fraud detection and motivate future work on dependence-aware fidelity assessment and hardware-aware evaluation under realistic NISQ constraints.

\section*{Acknowledgment}
This work was supported in part by the NYUAD Center for Quantum and Topological Systems (CQTS), funded by Tamkeen under the NYUAD Research Institute grant CG008, and the Center for Cyber Security (CCS), funded by Tamkeen under the NYUAD Research Institute Award G1104.

\bibliographystyle{IEEEtran}

\bibliography{refs}

\end{document}